\documentclass{article} %
\usepackage[submission]{colm2025_conference}

\usepackage{ulem}
\usepackage{comment}

\usepackage{microtype}
\usepackage{hyperref}
\usepackage{url}
\usepackage{booktabs}

\usepackage{lineno}

\usepackage{color,soul}

\usepackage{verbatim}

\definecolor{darkblue}{rgb}{0, 0, 0.5}
\hypersetup{colorlinks=true, citecolor=darkblue, linkcolor=darkblue, urlcolor=darkblue}

\usepackage{graphicx}

\usepackage{makecell}

\newenvironment{nohyphens}{%
  \par
  \hyphenpenalty=10000
  \exhyphenpenalty=10000
  \sloppy %
}{\par}

\usepackage{fancyvrb}

\title{LLM Library Learning Fails: A LEGO-Prover Case Study}

\colmsubmissionfalse
\colmpreprinttrue

\author{Ian Berlot-Attwell$^{\heartsuit\spadesuit}$, Frank Rudzicz$^{\spadesuit\clubsuit}$, Xujie Si$^{\heartsuit\spadesuit}$ \\
$^{\heartsuit}$University of Toronto, $^{\spadesuit}$Vector Institute, $^\clubsuit$Dalhousie University\\
Correspondence: \texttt{ianberlot@cs.toronto.edu} \\
}

\begin{document}

\newcommand{\hlrm}[1]{{\sethlcolor{red}\hl{#1}}}
\newcommand{\rvs}{\textit{request db}}
\newcommand{\prover}{\textsc{Prover}}
\newcommand{\evolver}{\textsc{Evolver}}
\newcommand{\lp}{LEGO-Prover}
\newcommand{\dsp}{Draft-Sketch-Prove}
\newcommand{\subgoal}{Subgoal-Prover}

\ifcolmsubmission
\linenumbers
\fi

\maketitle

\begin{abstract}
Recent advancements in the coding, reasoning, and tool-using abilities of LLMs have spurred interest in library learning (i.e., online learning through the creation, storage, and retrieval of reusable and composable functions, knowledge, checklists, or lemmas). Such systems often promise improved task performance through the automatic creation of broadly applicable tools, as well as superior computational performance through the caching of reasoning (i.e., the storage of generated tools). However, we find strong reasons to be skeptical. We perform an in-depth examination of one such system, \lp, which purports to learn reusable lemmas for mathematical reasoning. We find no evidence of the direct reuse of learned lemmas, and find evidence against the soft reuse of learned lemmas (i.e., reuse by modifying relevant examples). Crucially, we find that \lp\ does \textit{not} in fact improve over the simple baseline of prompting the model -- the improvements in task accuracy vanish once computational cost is accounted for. %
Our findings suggest that serious misconceptions exist as to the effectiveness of these techniques, that a serious re-examination of the state of LLM-based library learning is required, and that we require much stronger standards for evaluation including behavioural analysis and ensuring that an equal computational budget is used for baselines.\footnote{We release our code on GitHub: \url{https://github.com/ikb-a/llm_lib_learning_fails}}

\end{abstract}

\section{Introduction}

Humans extend their capabilities, improve their reliability, and save mental effort through  creating broadly applicable, composable, and reusable tools and knowledge. Whether software libraries, standard operating procedures, or mathematical lemmas, these artifacts allow us to save effort, act predictably, act accurately, and more easily solve tasks.  Library learning systems aim to artificially reproduce this ability -- e.g., DreamCoder \citep{DBLP:conf/pldi/EllisWNSMHCST21} can learn to write code for complex text editing and LOGO drawing tasks by building up a library of increasingly complex composable functions. 

Recent advances in the coding, reasoning, and tool-use abilities of LLMs has raised the possibility of performing library learning with LLMs. Coding ability is required to create reusable and composable code, and tool-use ability is needed to leverage these self-created tools. Indeed, within the last year, several works have leveraged in-context learning to perform library learning -- prompting the LLM to generate and apply broadly applicable helper tools. The primary benefit claimed is typically improved performance \citep{nguyen2024dynasaur}, though the proposed reasons for these gains vary. %
Compared to directly generating task-specific code, \citet{DBLP:conf/icml/WangNF24} and \citet{DBLP:conf/iclr/YuanC000J24} hypothesize that library learning encourages simpler and more reliable functions. Similarly, \citet{DBLP:conf/emnlp/QianH0Q0J23} suggest that performance improves because the task is broken down into separate abstract reasoning and instantiation tasks (i.e., creating a generic tool before calling it with the specifics of the give problem). \citet{lego} propose that their system works through the accumulation of reusable relevant reference examples.
These systems may also claim reduced costs through  caching  these created tools \citep{cai2024large}.

In this work, we investigate the internal behaviour and black box performance of one such system: \lp\ \citep{lego}. \lp\ (i.e., LP in tables) is a LLM-based library learning system built for the autoformalization of theorems, i.e., converting natural language proofs into formal theorems verifiable by the Isabelle theorem prover \citep{DBLP:books/sp/Paulson94}. Specifically, \citet{lego} reported SotA performance at the time of publication, and stated that ``\lp\ enables LLMs to utilize existing skills retrieved from the library'' and  ``[m]odular and reusable skills are constantly added to the library to enable tackling increasingly intricate mathematical problems.'' Based on these claims, we aim to answer two important questions. First: Does \lp\  exhibit library learning behaviours? Second: Does \lp\ improve performance?

Our contributions are as follows. (1) For a variety of LLMs, we demonstrate that the learned library is not directly reused. (2) For a variety of LLMs, we provide evidence that the learned library does not serve as a useful reference for solving related tasks. (3) We demonstrate that \lp\ does not outperform repeatedly prompting the model after accounting for computational budget. Together, these points suggests that a serious re-examination of LLM-based library learning is required. (4) Based on our findings, we make concrete proposals for improving the development and evaluation of library learning systems.

\section{Background}

\lp\ was designed to convert natural language proofs into verifiable formal proofs. It does so through two families of concurrent processes that interact via a shared database: the \prover\ and \evolver. 

The \prover\ iterates over pending tasks: For each task, it first decomposes the natural language proof into requests for potentially useful lemmas (in the form of Isabelle lemma statements without proofs) and then rewords the natural language proof as a step-by-step proof, still in English. This step-by-step proof is then augmented with retrieved lemmas and converted into a formal proof which is either verified or rejected by Isabelle. Simultaneously, the \evolver\ attempts to prove the requested potentially useful lemmas, adding these to the shared database. The \evolver\ also attempts to ``evolve'' existing lemmas to make them more useful for the solving of retrieved pending tasks or lemma requests. The \prover\ cycles through a pool of unsolved problems; the maximum number of attempts per problem is a hyperparameter. In both the \prover\ and \evolver, all steps are performed via in-context learning with examples. Retrieval is performed via Ada embeddings.

We use \dsp\ \citep{jiang2023draft} as a baseline, as done by \citet{lego}. \dsp\ (i.e., DSP in tables) uses in-context learning to directly convert a natural language proof into an Isabelle proof, applying post-processing with heuristics to fill or correct gaps in the proof. In our experiments we use \citet{lego}'s post-processor and wrapper of the Isabelle verifier, increasing the timeout to 120s as used by \dsp\ -- this wrapper was derived from \dsp\ and applies a superset of the original \dsp\ heuristics for correcting generated proofs.

Note that the original implementation of \dsp\ generated all proof attempts before performing verification. For fairer comparison and reduced cost we modify \dsp\ to alternate between proof generation and evaluation as \lp\ does, i.e., once a problem is solved by \dsp\ we cease generating proofs for this task.

\section{Evaluating \lp\ Under Various LLMs}

We begin by evaluating \lp\ and \dsp\ on the miniF2F test set using three different LLMs (see Table \ref{tab:eval_by_attempts}). Following \citet{lego} we run all models for a fixed number of prover attempts. We improve upon the original experimental procedure by performing three trials and reporting the standard deviation. We detail models and hyperparameters in Appendix \ref{sec:hyperparam}.

Note that our experiments on the full miniF2F test set use gpt-4o-mini as the underlying LLM instead of ChatGPT-3.5-Turbo as used by \cite{lego}. This model was selected due to its superior cost effectiveness %
and improved reasoning, coding, and mathematics proficiency  \citep{openai4omini}. We find that our \dsp\ performance of 35.9 $\pm$ 0.9\% using 4o-mini is close to the 38.5\% GPT-3.5-Turbo accuracy reported by \citet{lego}. This makes the discrepancy with \citet{lego}'s reported 45.5\% (or 50\% using human proofs) \lp\ accuracy all the more striking. We investigated whether \lp\ may be particularly sensitive to prompt, but found no such evidence (see Appendix \ref{sec:stability} for details). 

With the exception of gpt-4o-mini, these results \textit{appear} to confirm the findings of \citet{lego}: \lp\ outperforms \dsp\ which suggests a benefit from library learning (at the very least under more powerful models).

\renewcommand*{\thefootnote}{\fnsymbol{footnote}}

\begin{table}[]
    \centering
    
{\small
\begin{tabular}{lllllll}
\hline
     & gpt-4o-mini\footnotemark[1]   & gpt-4o\footnotemark[2]   & gpt-4o\footnotemark[3]   & gpt-4o\footnotemark[4]   & o3-mini\footnotemark[3]   & o3-mini\footnotemark[4]   \\
\hline
 DSP & \textbf{35.9} $\pm$ \textbf{0.9\%}              & 23.1 $\pm$ 1.2\%         & 25.0 $\pm$ 0.0\%         & 25.0 $\pm$ 0.0\%         & 19.4 $\pm$ 2.4\%          & 19.4 $\pm$ 4.8\%          \\
 LP  & 25.8 $\pm$ 1.6\%              & \textbf{29.3} $\pm$ \textbf{3.1\%}         & \textbf{29.2} $\pm$ \textbf{0.0\%}         & \textbf{33.3} $\pm$ \textbf{8.3\%}         & \textbf{27.8} $\pm$ \textbf{4.8\%}          & \textbf{36.1} $\pm$ \textbf{4.8\%}          \\
\hline
\end{tabular}
}
{\footnotesize{\footnotemark[1]~100~attempts, miniF2F test set} \hspace{2.8cm}
\footnotesize{\footnotemark[2] 50 attempts, 20\% of miniF2F test set}}\\
\footnotesize{\footnotemark[3] 50 attempts, 10\% of miniF2F test set} \hspace{2cm}
\footnotesize{\footnotemark[4] 50 attempts, 5\% of miniF2F test set}\\
    \caption{Evaluation of \lp\ and \dsp\ for a fixed number of prover attempts on various subsets of the miniF2F test set. Due to the cost, we reduce the number of attempts and evaluate more costly models on a subset of the test set. Consequently, \lp\ and \dsp\ are comparable between rows, but not all columns can be directly compared. Standard deviations over 3 trials are reported.}
    \label{tab:eval_by_attempts}
\end{table}

\renewcommand*{\thefootnote}{\arabic{footnote}}

\section{A behavioural analysis of \lp}

Assuming there is an improvement, what aspect of \lp\ is responsible? The most obvious explanation is that \lp\ develops a library of useful lemmas. These may be broadly applicable and useful results that can be \textbf{directly reused} in the problem (as a mathematician might do when directly applying a theorem). Alternatively, these may be useful exemplars that are \textbf{softly reused} (i.e., fragments reproduced with minor edits). This behaviour is akin, e.g., to forking a code repository as the starting point for a new project.

This explanation hypothesizes two key behaviours: First, that the \lp\ can directly or indirectly leverage relevant lemmas in its proof writing. i.e., the \lp\ can \textbf{use} relevant lemmas. Second, that the library contains \textit{general} lemmas that are both relevant to and leveraged in several tasks. I.e., the \lp\ can \textbf{reuse} lemmas.

In the following sections, we examine both direct and indirect reuse, finding evidence for use but against both soft and direct reuse.

\subsection{Quantifying Soft Use}\label{sec:softusescore}

We quantify the soft use between a given input lemma and the \prover's solution as the minimum proportion of space-separated tokens in the lemma that must be deleted or substituted for the lemma to appear within the solution. 
We obtain this score by first computing the Levenshtein distance \citep{Levenshtein_SPD66} (i.e., the minimum number of deletions, replacements, or insertions required to convert one string to another) assigning a weight of 0 to insertions. We allow for an arbitrary number of insertions, as a used lemma is likely to be only a fragment of the final proof. This modified Levenshtein distance thus has a value between 0 (all tokens in the lemma also appear in the solution and in the same order) and, $N$, the number of tokens in the lemma (no tokens in the lemma appear in the solution). We divide this value by the length of the lemma to produce a normalized score for soft use. See Appendix \ref{app:soft-use-examples} for examples of various lemmas and their soft use scores.

\subsection{Does the \prover\ use the input lemmas?}\label{sec:use}

\begin{table}
  \centering
  \begin{tabular}{lrrr}
    \toprule
    Model & \% of Test Set & Verbatim Use & Name Use \\
    \midrule
    GPT-4o & 20\%  & 2/43 &  9/43    \\
    GPT-4o & 10\%  & 4/21 & 6/21     \\
    o3-mini & 10\% & 0/20 & 5/20     \\
    4o-mini & 100\%  & 13/189 & 47/189     \\
    \bottomrule
  \end{tabular}
    \caption{Proportion of found solutions in which a retrieved lemma or its name appears in the solution. We observe that \lp's \prover\ can reproduce the retrieved input lemmas, however Table \ref{tab:lp_reuse} reveals that these lemmas are not reused across tasks. We report sums over 3 runs. Note that we run some models on random subsets of the miniF2F test set due to cost limitations.}
  \label{tab:lp_use}
\end{table}

A prerequisite for tool reuse is tool use: a tool cannot be used several times if one cannot use it at all. 

To test for direct use, we apply the methodology by \citet{berlot-attwell2024library}: for every retrieved lemma provided to the \prover\ we directly check for the presence of the lemma or the lemma's name in the \prover's output (the latter being a generous way of accounting for changes such as whitespace or comments that do not impact semantics). We find that \lp\ is capable of direct use with gpt-4o and 4o-mini (i.e., directly copying an input lemma into the final proof -- see Table \ref{tab:lp_use}). While we do not find direct use with o3-mini, a manual inspection of the instances of name use finds two solutions where the lemmas were copied from the input with some modifications (i.e., soft use).

We also find evidence for indirect use. For any given threshold of the soft use score, we can calculate the proportion of input lemmas with a soft-use score of at least the threshold. This defines a survival function with the threshold as the independent variable. The greater the extent of soft use exhibited by the system, the slower we expect this curve to decay. For systems that perform direct use, this curve does not reach zero as the threshold approaches one. Instead, the lower bound is the proportion of lemmas that are directly used. To analyze soft-use behaviour, we compare the survival function of retrieved lemmas against that of other baseline populations of lemmas. The first baseline is a pool of non-retrieved lemmas produced by \lp\ (i.e., lemmas written by the same LLM for the same tasks but \textit{not} retrieved and provided to the \prover). These non-retrieved lemmas represent semantically and stylistically similar lemmas. The second baseline is a pool of unrelated human-written lemmas extracted from the 2019 Archive of Formal Proofs (AFP)\footnote{\url{https://www.isa-afp.org/}; see Appendix \ref{sec:hyperparam} for details.}. 

These survival curves are plotted in the top row of Figure \ref{fig:lemma_surv}. As expected, we see that the survival curve of unrelated AFP lemmas drops fastest and rapidly reaches zero (demonstrating that our soft use metric is capturing relevant information). At high similarity thresholds, the retrieved lemmas maintain a larger population than the non-retrieved lemmas, thus providing evidence for soft use. Note that o3-mini is consistent with the trend, but has noisier measurements, as it was evaluated on a smaller dataset due to cost limitations.

\subsection{Does the \lp\  learn a library?}

The ability to use tools is a necessary but insufficient condition for library learning. Unlike mere task decomposition, library learning must create \textit{reusable tools}. We must therefore inspect the reuse behaviour of \lp. A lemma is said to be reused $n$ times by the \prover\ if it is used in $n+1$ proofs (i.e., the initial use, followed by $n$ reuses). 

\paragraph {Direct Reuse}

 We find no evidence for the direct reuse of lemmas (see Table \ref{tab:lp_reuse}). We find only one instance of name reuse in 189 successful proofs generated by 4o-mini. Manual inspection reveals that this reuse does not appear in the final verified proof as the reuse was part of an erroneous proof step generated by the LLM that was replaced by the human-written proof correction heuristics.

\begin{table}
  \centering
  \begin{tabular}{llrrllll}
    \toprule
    \multicolumn{4}{c}{} & \multicolumn{2}{c}{Verbatim reused} & \multicolumn{2}{c}{Name reused} \\
    \cmidrule(r){5-6}
    \cmidrule(r){7-8}
    Model & \% of Test Set & \makecell{Successful \\ Attempts}  & \makecell{Lemmas in\\ Prompts} & 1 & 2+ & 1 & 2+ \\
    \midrule
    GPT-4o &20\% & 43/5537  & 121  & 0 & 0 & 0 & 0     \\
    GPT-4o & 10\% & 21/2727  & 64 & 0 & 0 & 0 & 0     \\
    o3-mini & 10\% & 20/1825 & 75 & 0 & 0 & 0 & 0     \\
    4o-mini & 100\% & 189/57419 & 583 & 0 & 0 & 1 & 0     \\
    \bottomrule
  \end{tabular}
    \caption{Lemma reuse in successful \prover\ attempts. For various models and random subsets of the miniF2F test set, we report the proportion of successful \prover\ attempts, the number of unique retrieved lemmas occurring in the \prover's input prompts, the number of lemmas reused verbatim once, or more than once, and the number of lemmas whose \textit{name} is reused once, or more than once. A lemma is reused $N$ times if it appears in $N+1$ solutions (i.e., the initial use, and then $N$ reuses). We sum all figures over 3 runs.}
  \label{tab:lp_reuse}
\end{table}

\paragraph{Soft Reuse}

\begin{figure}
    \centering
    \includegraphics[width=1\linewidth]{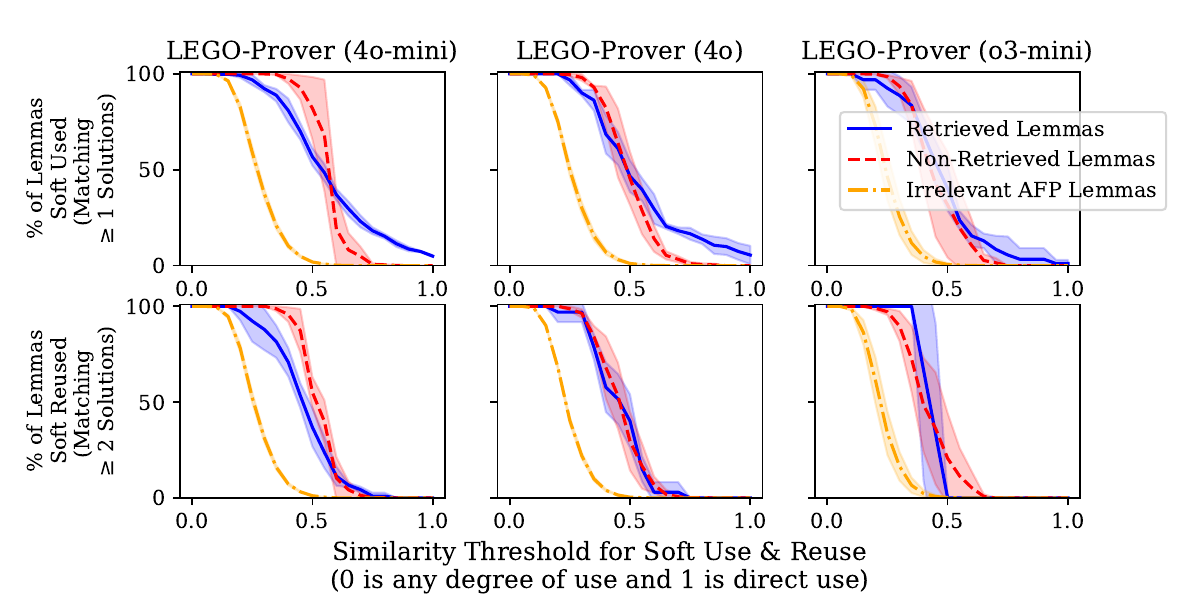}
    \caption{%
    These plots demonstrate that \lp\ exhibits soft use, but not reuse, of lemmas. The top row concerns use. It plots the percentage of lemmas meeting a given the threshold for soft use in at least one solution (see Section \ref{sec:softusescore} for the definition of this score). The lemmas available to the \prover\ are shown in the solid blue line, these are the only lemmas for which use is possible. As expected, in the top row this line remains the highest for large values of the threshold, surpassing the non-retrieved lemmas (red dashed line). However, in the bottom row we consider the proportion of lemmas meeting the soft use threshold in \textit{two} tasks (i.e., reused by \lp). Unlike in the top row, there is no significant difference between the lines -- suggesting retrieved lemmas are only meaningful in a single task (consequently, used, not reused). 
    See Section \ref{sec:use} for a detailed explanation.}
    \label{fig:lemma_surv}
\end{figure}

While these results bode poorly, the original authors %
claim that ``[m]any skills cannot be directly reused but are very helpful as reference examples for formalizing the main problem.'' Therefore, to study \lp's soft reuse behaviour, we repeat our survival curves from Section \ref{sec:use} but instead plot the proportion of lemmas used in at least two solutions generated by the LLM (see Figure \ref{fig:lemma_surv}, bottom row). In sharp contrast to our single-use findings there is no significant difference between the retrieved and non-retrieved lemmas. This suggests that \lp\ learns overly-specific lemmas that are not reused in other problems.

\begin{figure}
    \centering
    \includegraphics[width=1\linewidth]{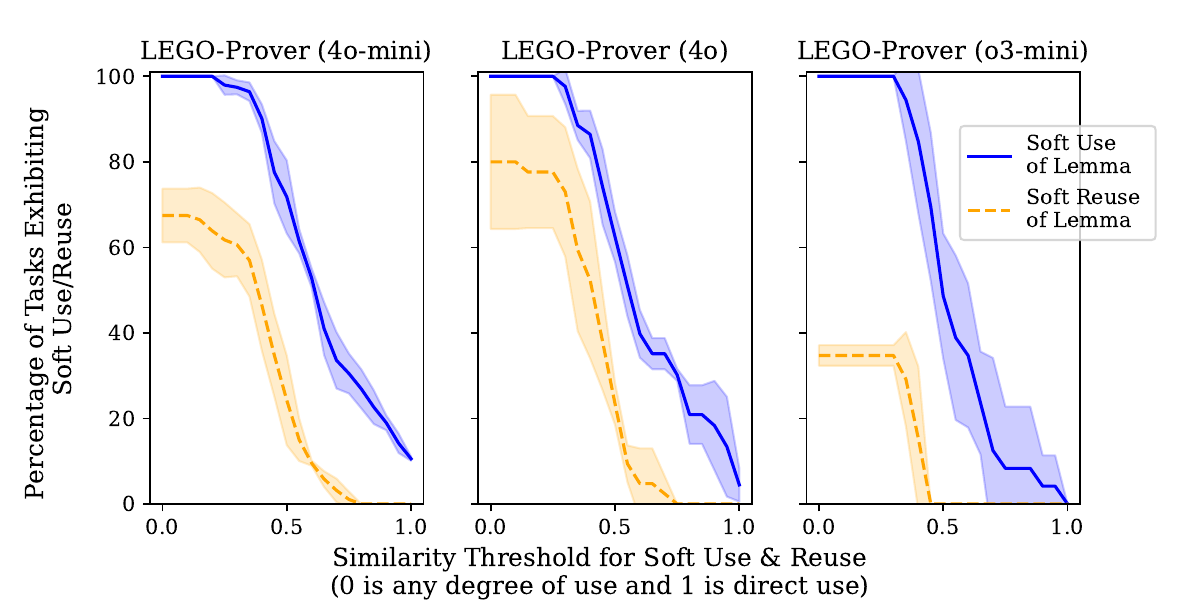}
    \caption{We define a normalized score for the degree to which a lemma is used in a proof (i.e., soft use). Using this score and a threshold, we can determine whether \lp's solution to a task soft-uses the retrieved lemmas provided in the \prover's context. In blue we plot the percentage of solutions meeting the desired threshold for soft use. In orange, we plot the percentage of solutions meeting the threshold in a lemma that is also used at the same level in at least one other solution. We observe that the lines for reuse rapidly reach zero for even moderate similarity thresholds wheras the lines for use remain relatively high. This suggests that lemmas are used by the \prover\ in generating solutions, but that there tends to be only one task that exploits a given lemma. }
    \label{fig:task_surv}
\end{figure}

As an alternative analysis, we also plot the survival function for the proportion of solved tasks exhibiting soft use, or soft reuse, at a given threshold (See Figure \ref{fig:task_surv}). We note that the soft reuse curves are within one standard deviation of 0\% of tasks by the 70\% soft-reuse threshold, and reach zero while 30-40\% of tasks continue exhibiting soft use. This suggests that soft reuse is not occurring. Note that these curves also reveal an upper bound on the proportion of tasks that can benefit from lemma reuse, given the created library. Unlike our previous analyses, this bound arises from the \textit{retriever} that selects lemmas from the library. Specifically, some tasks do not retrieve any lemmas that are also retrieved by another task. Therefore, it is impossible for these tasks to exhibit \prover\ reuse. Consequently, at a soft use threshold of zero (i.e., all retrieved lemmas are considered to be soft used) the task survival functions for reuse still attains a value of less than 100\% -- this value is an upper bound on the proportion of tasks that can benefit from reuse in the \prover, assuming that the retriever and library remain unchanged.

\section{What drives \lp's superior miniF2F performance?}

Table \ref{tab:eval_by_attempts} \textit{appears} to demonstrate that \lp\ outperforms \dsp. \citet{lego}, naturally, claim that library learning is the mechanism for this improvement. \textit{However}, we have shown that \lp\ does not reuse the learned library. As such, we claim that the improvements instead arise from increased compute. The original data compares \dsp\ and \lp\ based on a fixed number of \prover\ attempts. However, this does not account for the compute used by the \evolver, which runs in parallel to the \prover. The \evolver's compute is significant, as the default hyperparameters create twice as many \evolver\ processes than \prover\ processes. Furthermore, the \prover\ itself is more computationally intensive, as it performs a 2-stage process of decomposition and formalization. Using cost as a proxy for GPU compute, we estimate that each \lp\ attempt uses approximately 6 times more GPU compute than \dsp, increasing to 14 times with o3-mini (see Table \ref{tab:compute_ratios}). Adjusting for the GPU budget, we find that \dsp\ matches or outperforms \lp\ (see Table \ref{tab:eval_by_cost}).

\begin{table}[]
    \centering
\begin{tabular}{lllllll}
\hline
  & gpt-4o-mini   & gpt-4o  & o3-mini   \\
\hline
 GPU Budget Ratio & 5.84 & 5.94 & 14.23   \\
\hline
\end{tabular}
    \caption{We find that \lp\ uses significantly more GPU resources that \dsp\ per attempt, and calculate the ratio of the costs under a fixed number of attempts. This ratio grows larger for o3-mini as \lp\ uses significantly more output tokens per input token. We speculate that either lemma creation %
    or reasoning about the use of lemmas %
    require a higher proportion of reasoning tokens than directly solving the task.}
    \label{tab:compute_ratios}
\end{table}

\renewcommand*{\thefootnote}{\fnsymbol{footnote}}

\begin{table}[]
    \centering

{\small
\begin{tabular}{lllllll}
\hline
             & gpt-4o-mini\footnotemark[1]   & gpt-4o\footnotemark[2]   & gpt-4o\footnotemark[3]   & gpt-4o\footnotemark[4]   & o3-mini\footnotemark[3]   & o3-mini\footnotemark[4]   \\
\hline
 DSP         & \textbf{35.9} $\pm$ \textbf{0.9\%}              & 23.1 $\pm$ 1.2\%         & \textbf{25.0} $\pm$ \textbf{0.0\%}         & \textbf{25.0} $\pm$ \textbf{0.0\%}         & \textbf{19.4} $\pm$ \textbf{2.4\%}          & \textbf{19.4} $\pm$ \textbf{4.8\%}          \\
 LP          & 17.5 $\pm$ 0.9\%              & \textbf{23.8} $\pm$ \textbf{1.2\%}         & 22.2 $\pm$ 2.4\%         & 22.2 $\pm$ 4.8\%         & 15.3 $\pm$ 2.4\%          & \textbf{19.4} $\pm$ \textbf{9.6\%}          \\
 LP attempts & 17                            & 9                        & 8                        & 6                        & 4                         & 3                         \\
\hline
\end{tabular}
}
\noindent\footnotesize{\footnotemark[1]~100~DSP attempts, miniF2F test set} \hspace{2.8cm}
\footnotesize{\footnotemark[2] 50 DSP attempts, 20\% of miniF2F test set}\\
\footnotesize{\footnotemark[3] 50 DSP attempts, 10\% of miniF2F test set}\hspace{2cm}
\footnotesize{\footnotemark[4] 50 DSP attempts, 5\% of miniF2F test set}\\

    \caption{A modification of Table \ref{tab:eval_by_attempts} that limits the number of \lp\ attempts to match \dsp's compute budget. Note that \dsp\ consistently outperforms \lp\ or is within one standard deviation of \lp's performance. Average accuracy and standard deviations are again reported over 3 trials.}
    \label{tab:eval_by_cost}
\end{table}

\renewcommand*{\thefootnote}{\arabic{footnote}}

\begin{figure}
    \centering
    \includegraphics[width=1\linewidth]{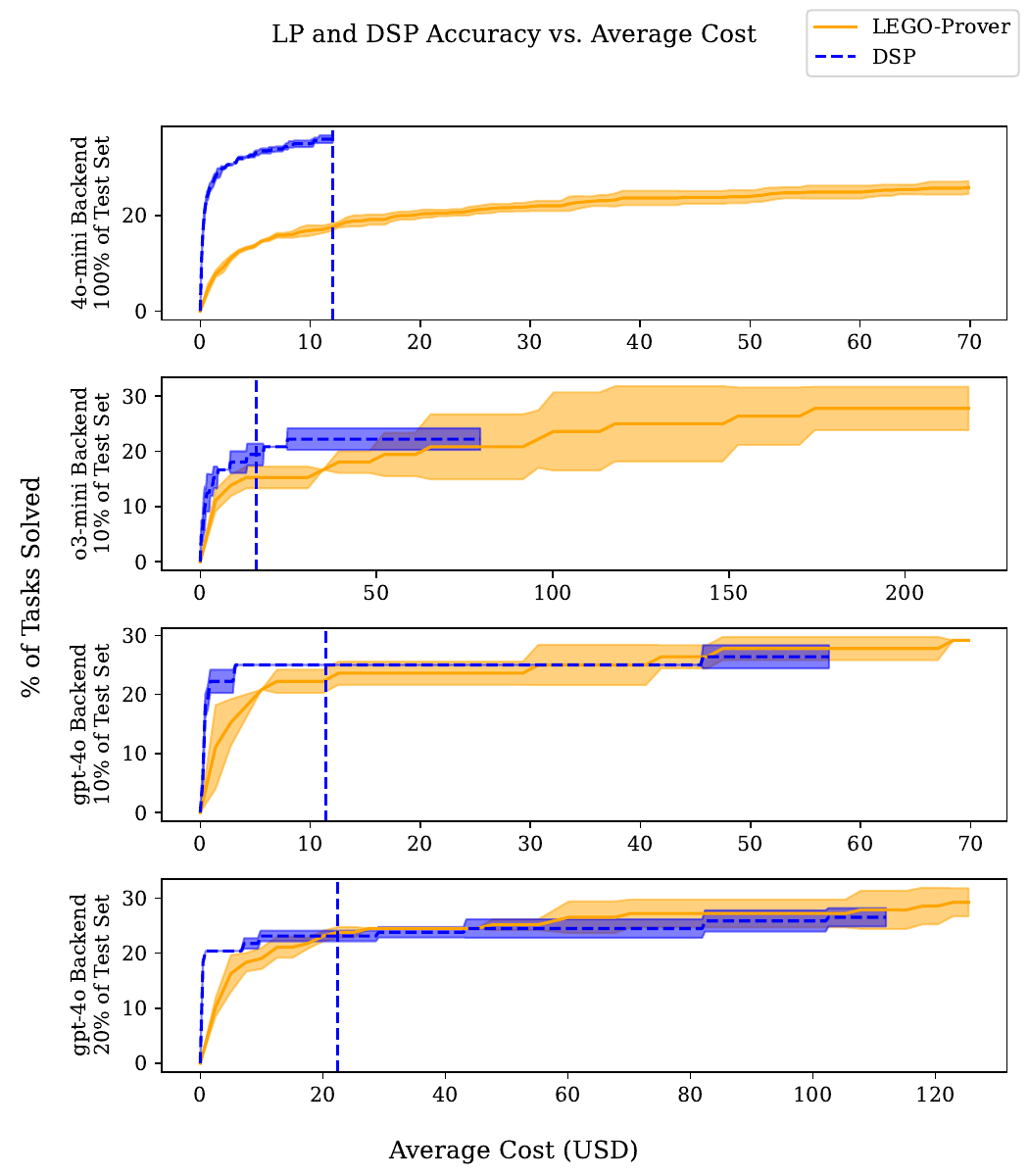}
    \caption{Average \lp\ and \dsp\ performance compared by average cost in USD.  The shaded region is 1 standard deviation based on 3 trials.  \dsp\ has strongly superior performance using 4o-mini, and comparable performance on gpt-4o and o3-mini. The vertical line represents the number of \dsp\ prover attempts equal to \lp's. For reasons of cost o3-mini was run on a random 10\% subset of the test set and gpt-4o was run on the same random 10\% subset as well as a 20\% subset. Note that as the cost of o3-mini was over twice our predicted cost we were unable to run \dsp\ for a comparable budget. See Appendix \ref{app:cost_plot_creation} for the details of how this plot was generated. %
    }
    \label{fig:cost}
\end{figure}

However, this analysis it limited -- it is possible that \dsp\ initially matches the performance before plateauing ahead of \lp, or that \lp\ requires a certain number of iterations for the library to reach a critical mass. Therefore, we run a followup experiment in which we extend our \dsp\ runs, increasing the number of iterations to approximately match the GPU resources used using cost as as proxy. Doing so, we find that \dsp\ remains within one standard deviation or outperforms \lp\ throughout the evaluation (see Figure \ref{fig:cost}).

\section{Related Work}

Within the specific area of LLM-based library learning, there has been growing evidence that these systems do not perform direct reuse. \citet{berlot-attwell2024library} previously studied \lp\ and TroVE \citep{DBLP:conf/icml/WangNF24}, finding negligible direct reuse within these systems and instead proposing that the performance gains are due to self-correction and indirect reuse. Similarly, \citet{nguyen2024dynasaur} reported that their library learning DynaSaur system achieved SoTA results on the GAIA agent benchmark \citep{gaia}, but exhibited low direct reuse of the learned libraries. They did not  speculate on the implications, however, beyond that it may be addressed via curriculum learning. Our work (1) provides further evidence against direct reuse %
(to the best of our knowledge we are the first to evaluate direct reuse in a reasoning model),
(2) advances our understanding by providing new evidence against the soft reuse hypothesis, and (3) provides a simple compute-based hypothesis as to why these systems outperform their baselines -- they increase the test-time compute per attempt in a way that is not accounted for in typical evaluation methods.

In the related domain of Python code generation, \citet{olausson2024repair} studied the relative efficacy of self-correction as compared to increased sampling  for the APPS \citep{DBLP:conf/nips/HendrycksBKMAGB21} and HumanEval \citep{chen2021codex} tasks. Unlike our results, they found a moderate effect when accounting for compute. They hypothesized that ``current models are held back by their inability to reliably produce accurate and useful feedback on why the code is wrong.'' Given the known success of tool-using LLMs (e.g., ViperGPT \citep{surismenon2023vipergpt} achieved SotA performance on vision-language tasks using a code generating LLM conditioned on a human provided Python library for vision processing), we suspect that there may be a similar bottleneck in library learning. I.e., these systems may be held back by an inability to reliably produce general theorems and would improve in performance if given human-created libraries.

In the domain of agentic AI,  \citet{DBLP:journals/corr/abs-2407-01502} call for the use of cost as a metric and claim that planning, reflection, and debugging offer no improvement over trivial baselines on the HumanEval \citep{chen2021codex} task. Our work provides similar findings in the area of autoformalization when using the techniques of library learning. This is interesting, as autoformalization is closer to a translation task rather than generation (i.e., converting a natural language proof to code) and, 
unlike Python programming, we have access to a strong verifier that can confirm the correctness of our reasoning (i.e., unlike self-correction that may introduce errors by fixing hallucinatory errors, all intermediary lemmas are verified by Isabelle). Despite these advantages, we do not to find an improvement in accuracy. Furthermore, given our focus on library learning specifically, we also go into the details of whether the system internally behaves as expected, as opposed to studying the general trends of cost and accuracy. In doing so, we find evidence that these techniques are ineffective and fail to operate as believed, suggesting that there are fundamental flaws in these specific approaches.  %

\section{Conclusions}

We have gained a number of insights into the \lp\ system. Firstly, \lp\ does not exhibit reuse behaviour -- therefore, it does not perform library learning. Secondly, \lp\ does not improve performance over a simple prompting baseline once computational cost is accounted for.

These findings, particularly with respect to accounting for computational budget, become wide-reaching when considered in the broader literature of LLM library learning. Our work combines with other emerging evidence to suggest serious flaws in the assessment of LLM-based library learning. It was believed that LLM library learning works, as it improves performance -- but our findings on the importance of accounting for computational cost cast this into question. We are unaware of \textit{any} LLM-based library learning work that controls for compute in its evaluation. It is possible that the observed benefits of library learning are simply due to a hidden form of inference-time scaling. This calls into the question the efficacy of all these works -- and we cannot hope to make progress without resolving this problem. This would also go towards explaining other findings: previous work has demonstrated that the library learning system TroVE does not outperform library-free ablations \citep{berlot-attwell2024library} -- whereas our work can explain this finding as \citet{berlot-attwell2024library}'s ablation implicitly matched computational cost. Our work can also explain apparent paradoxes -- the DynaSaur library learning system reports SoTA performance and a lack of reuse behaviour \citep{nguyen2024dynasaur}. This contradiction is resolved if increased inference-time compute is responsible for the performance gains. Indeed, in their evaluation, they fixed all models to 20 steps. I.e., their evaluation did not account for the differences in computational cost per step between DynaSaur, direct prompting of gpt-4o, Sibyl \citep{DBLP:journals/corr/abs-2407-10718}, and HF Agent \citep{hfagent}.

These findings also call into question the argument that LLM library learners can reduce costs (at least in their current form). The caching of generated tools is moot if said tools are not reused, and our analysis suggests that they are not. Secondly, the savings from caching must exceed the cost of creating and selecting these tools -- we perform the first cost-controlled comparison of an LLM library learner that we are aware of and find no improvement over prompting with the same budget.

Looking to how we can build better systems -- from the perspective of accuracy, our work suggests that our effort may be best directed towards improving base models. From the perspective of library learning, our work calls into question the innate ability of LLMs to perform refactoring using in-context learning alone. Our suggestion would therefore be towards incorporating symbolic refactoring systems that are known to work (e.g., the LILO \citep{DBLP:conf/iclr/GrandWBOLTA24} system takes this approach). If we instead wish to use LLMs directly for library learning then, as context learning is not enough, it may be worthwhile to explore finetuning or RL (e.g., via GRPO \citep{DBLP:journals/corr/abs-2501-12948} with human-written reward functions, possibly based on compression). Critically, any such system must be evaluated thoroughly. Therefore, any system claiming to perform library learning should be evaluated with respect to both accuracy \textbf{accounting for computational budget}, and the system's behaviour. The extent of reuse must be measured, be it direct or soft reuse. In service of this goal, we  also suggest the construction of a synthetic diagnostic dataset based on a known ground truth library.

\section*{Acknowledgments}

Resources used in preparing this research were provided, in part, by the Province of Ontario, the Government of Canada through CIFAR, and companies sponsoring the Vector
Institute \url{www.vectorinstitute.ai/partnerships/}.
Generous support was also provided by the Microsoft Accelerating Foundation Models Research (AFMR) program.

We would also like to thank Francois Roewer-Despres and Haokun Liu for their time and conversations that helped in various ways to shape and improve this work. We would also like to thank Steven Zhong for his technical support, without which this project would not have been possible.

\bibliography{colm2025_conference}
\bibliographystyle{colm2025_conference}

\appendix

\section{Technical Details \& Hyperparameters}\label{sec:hyperparam}

In our experiments we use gpt-4o version 2024-08-06, o3-mini version 2025-01-31, 4o-mini version 2024-07-18, and text-embedding-ada-002.

\begin{table}[t]
\begin{center}
\begin{tabular}{ll}
\toprule
\multicolumn{1}{c}{\bf \lp\ Hyperparameter}  &\multicolumn{1}{c}{\bf Value} \\
\midrule
Temperature         & 0.7, \underline{(N/A for o3-mini)} \\
\# \prover\ Processes             & 3 \\
\# \evolver\ Processes              & 8 \\
\# Attempts & 100 for full miniF2F test set, else 50 \\
\evolver\ maximum tokens decoded & 1024, (\underline{8092 for o3-mini}) \\
\prover\ maximum tokens decoded & 2000, (\underline{8092 for o3-mini}) \\
\bottomrule
\end{tabular}
\end{center}
\caption{\lp\ hyperparameters. Values changed from \citet{lego} are underlined.}\label{tab:lp_hyper}
\end{table}

\begin{table}[t]
\begin{center}
\begin{tabular}{ll}
\toprule
\multicolumn{1}{c}{\bf \dsp\ Hyperparameter}  &\multicolumn{1}{c}{\bf Value} \\
\midrule
Temperature         & 0.0, \underline{(N/A for o3-mini)} \\
\# \ Processes             & 11\\
\# Attempts & 100 for full miniF2F test set, \underline{else 50} \\
Maximum tokens decoded & 2048, (\underline{8092 for o3-mini}) \\
\bottomrule
\end{tabular}
\end{center}
\caption{\dsp\ hyperparameters. Values changed from \citet{jiang2023draft} are underlined.}\label{tab:dsp_hyper}
\end{table}

\lp's hyperparameters are listed in Table \ref{tab:lp_hyper}; values changed from defaults are underlined. Note that \citet{lego} use 50 attempts for their ablation experiments. \dsp's hyperparameters are listed in Table \ref{tab:dsp_hyper}. Note that we modified \dsp\ to run 11 processes in parallel (with each launching its own Isabelle instance) so that the average resources per Isabelle instance is equal to \lp. The prompts used in our experiment are documented in Appendix \ref{sec:prompts}. All our experiments use Isabelle 2022.

We used fixed, random, non-overlapping subsets of 5\%, 10\% and 20\% of the miniF2F test set for our experiments. For reasons of cost, we only study system behaviour with human-written English proofs as inputs to the systems.

Our pool of unrelated Isabelle lemmas used in Figures \ref{fig:lemma_surv} and \ref{fig:task_surv} are extracted from the \texttt{2019-08-19} Archive of Formal Proofs\footnote{Available for download at the time of publication from \url{https://sourceforge.net/projects/afp/files/afp-Isabelle2019/}}. Specifically, we used all matches to the following Python3 regex: 

\begin{verbatim}
(?:lemma|theorem)\s+[^:]+?\s*:(?:[\s\S](?!lemma|theorem))+?qed    
\end{verbatim}

This regex matches lemmas and theorems with similar syntax to the in-context examples, though it may prematurely truncate some lemmas and theorems.

Each experiment is performed with a resource allocation of 180~GB of RAM and 50 Intel Broadwell CPU cores @2.095~GHz. These resources are shared among all Isabelle instances.

\section{Limitations}

Due to reasons of cost, we were unable to run o3-mini or gpt-4o on the full test set. o3-mini in particular was far more costly than expected As such, we could only afford to evaluate on a 10\% subset of the test set. Having said this, we do observe the same trends across all three of the LLMs studied, and we were able to evaluate gpt-4o-mini on the full test set.

Cost and a lack of access to the underlying model also raise barriers to the reproduction of this work. We are releasing all of our collected data in an effort to lower barriers, as well as releasing our code so that it may be adapted to other LLMs. This data also includes the token usage for all of our experiments, allowing for more precise estimates of cost, as well as the estimation of costs under other LLMs. We also used gpt-4o-mini for our largest experiments, reducing the cost of these experiments. 

In the interests of improved reproducibility we also attempted evaluation using \texttt{DeepSeek-R1-Distill-Qwen-1.5B},
\texttt{DeepSeek-R1-Distill-Qwen-7B}, \texttt{DeepSeek-R1-Distill-Llama-8B}, and \texttt{DeepSeek-R1-Distill-Llama-70B}. Unfortunately, were unable to get the \dsp\ baseline to work due as we found all of these models posses a strong bias for producing Lean 4 code. On the rare occasions when the models did produce Isabelle code, it was still produced within a \texttt{lean} code block -- based on this we suspect that lean theorem proving was used in the distillation data, and possibly also as one of the RL tasks used to originally train DeepSeek R1.

There are also limitations to the measurement of cost. Firstly, we could only measure the token usage of completed runs, therefore there may be slight distortions in Figure \ref{fig:cost} as it was necessary to plot based on the average cost per attempt. In reality, it will be the case (at least for \dsp) that the cost per attempt reduces slightly over time, as the pool of problems to solve shrinks and fewer problems must be tried per attempt. Secondly, \lp\ uses the \texttt{text-embedding-ada-002} model for the retrieval of lemmas, lemma requests, and pending tasks. For some \lp\ logs we were unable to log Ada usage. As such we chose to remove Ada from the cost for all of our plots -- in practice the cost of Ada was quite low and this omission reduces \lp's running costs, therefore we do not consider this to be an issue. 

Another limitation applies to our \lp\ evaluations with gpt-4o and o3-mini, as well as our prompt stability experiments in Appendix \ref{sec:stability}. We performed our experiments using the released \lp\ code base, however we realized after our experiments were completed that this code base forcibly initializes the \evolver\ with a database of all test queries. As such, the \lp\ evolver has access to information about other tasks. The impact of this on the \lp's performance is unclear: on one hand our intuition suggests that this may distract the \lp\ from the tasks at hand, hindering performance until new \prover requests take priority. On the other hand, we performed some informal experiments with \lp\ ablations using gpt-4o-mini and found that preventing the \evolver\ from retrieving these entries counter-intuitively slowed the rate at which problems were solved. Regardless, given the vast gap in cost (which persists in our 4o-mini experiments for which the \evolver\ was initialized with exactly the set of tasks to be solved), we do not believe that adjusting the \evolver's initialization will have a significant impact on our results.

Other limitations are due to our chosen soft use score. Firstly, our key assumption was that soft use would take a form similar to the modification of starter code. It remains possible that there are other forms of soft reuse (e.g., ones that involves substantial paraphrasing or merely imitate the general structure of the lemma) that our soft use score would not detect. Having said this, as the difference between the text and the proof increases, our intuition dictates that the likelihood of the introduction of an error would increase (as the model would be further distancing itself from the Isabelle-verified code), and thus the value of such reuse would decrease.  Secondly, as our metric does not penalize the insertion of new tokens, it almost certainly scores shorter lemmas higher on average. Having said this, we believed that this trade-off was acceptable, given the premise that the final proof would (and, indeed, in any working library learning system \textit{should}) contain more than just the lemma -- e.g., a proof may use the Pythagorean theorem, but it would do so in conjunction with other axioms, lemmas, and reasoning steps. 

One concern is that gpt-4o and o4-mini scores appear to be low on our random test subsets. We suspect that random test sets consist of harder problems. Specifically, we can observe in Tables \ref{tab:eval_by_attempts} \& \ref{tab:stability} that while \dsp\ reproduces \citet{lego}'s reported overall miniF2F test set accuracy, its performance drops by 15.9\% (absolute) on our random 5\% subset of the test set.

Another possible limitation follows from our observation that o3-mini has difficulties placing the solutions to lemma requests inside of markdown code blocks. Through manual prompt tuning we were able to improve the reliability but it remains imperfect -- it is possible that this behaviour reduces the performance of \lp\ by slowing the \evolver. Closely related, as neither \dsp\ nor \lp\ were designed for reasoning models, their prompts are suboptimal. OpenAI recommends that prompts be short, and not contain examples. Thus the options were to use suboptimal prompts, or to significantly modify the prompts of both models (which, due to the low temperature used by \dsp, could eliminate variations in \dsp\ prompt required to solicit diverse LLM responses). Ultimately we erred towards the side of caution and chose to make only the minimal changes required to correct any gross errors of behaviour in \dsp\ or \lp\ (e.g., formatting errors, producing proofs for example problems, etc...). Based on our prompt stability experiments (see Appendix \ref{sec:stability}) we believe that minor rephrasing of our final prompts would not seriously change the results, though it remains possible that o3-mini behaves significantly differently to gpt-4o-mini in this regard.

Another concern is that due to the changes made to our prompts the resulting prompts may be suboptimal. To address this possibility we study the prompt sensitivity of \dsp\ and \lp; the results are outlined in Appendix \ref{sec:stability}.

\subsection{Stability with Respect to Prompt} \label{sec:stability}

\begin{table}[t]
    \centering
\begin{tabular}{lllll}
\hline
 model   & val Baseline acc   & val Paraphrase acc   & Max Pot. Gain   & Max Pot. Loss   \\
\hline
 dsp     & $50.0\pm 0.0$\%    & $48.3\pm 3.7$\%      & 0.0\%           & 0.0\%           \\
 lp      & $38.3\pm 4.6$\%    & $38.3\pm 4.6$\%      & 0.0\%           & 0.0\%           \\
\hline
\end{tabular}
\begin{tabular}{lllll}
\hline
 model   & test Baseline acc   & test Paraphrase acc   & Max Pot. Gain   & Max Pot. Loss   \\
\hline
 dsp     & $16.7\pm 0.0$\%     & $20.0\pm 4.6$\%       & 16.7\%          & 0.0\%           \\
 lp      & $18.3\pm 3.7$\%     & $16.7\pm 5.9$\%       & 0.0\%           & 0.0\%           \\
\hline
\end{tabular}
    \caption{Average test accuracy of the models under both our baseline prompt and paraphrases of the prompt. We also report variability in the results, following the methodology for prompt sensitivity from \citet{wahle-etal-2024-paraphrase}. All experiments were run using 50 attempts, 5 runs, and on a 5\% split of the miniF2F validation and test sets respectively.}
    \label{tab:stability}
\end{table}

For both \dsp\  and \lp\ we were required to make minor modifications to the prompt to ensure that 4o-mini generated responses with the correct formatting (See Appendix \ref{sec:prompts} for the prompts and our modifications). To address the concern that our choice of prompt may be disadvantaging \lp, we generate five prompt paraphrases, and following \citet{wahle-etal-2024-paraphrase} study the relative stability under the paraphrased prompts versus the original prompt. We generate the prompt paraphrases using GPT3.5, as done by \citet{chatterjee-etal-2024-posix}.

\begin{figure}
    \centering
    \includegraphics[width=0.5\linewidth]{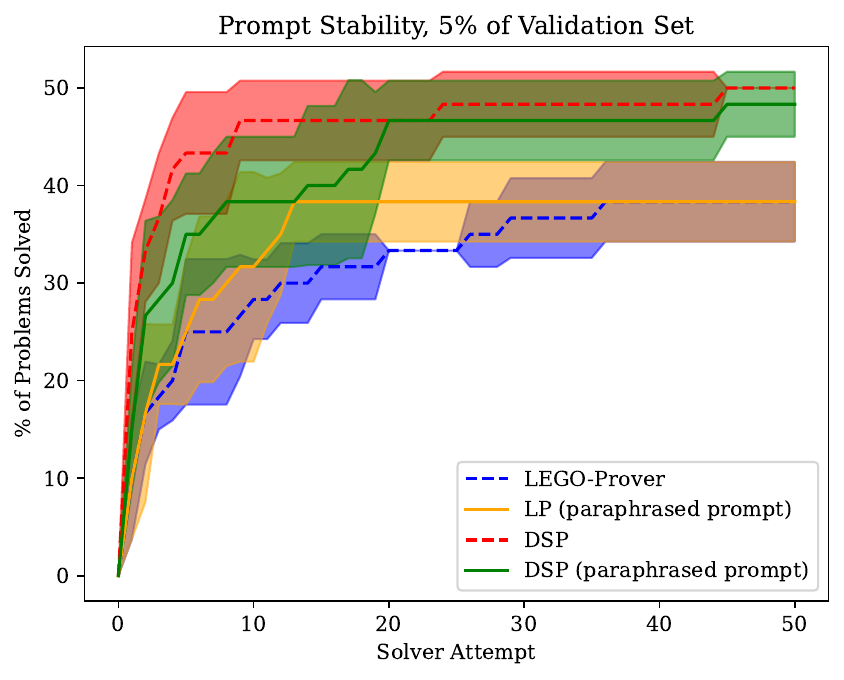}
    \caption{For \lp\ and \dsp\ we compare 5 runs with our default prompt versus 5 runs with different paraphrased prompts. We evaluate on a subset of the validation set. We find that the systems are fairly robust to minor changes in wording -- performance is typically within 1 standard deviation. These results are for a 5\% subset of the validation split.}
    \label{fig:stab-val}
\end{figure}

\begin{figure}
    \centering
    \includegraphics[width=0.5\linewidth]{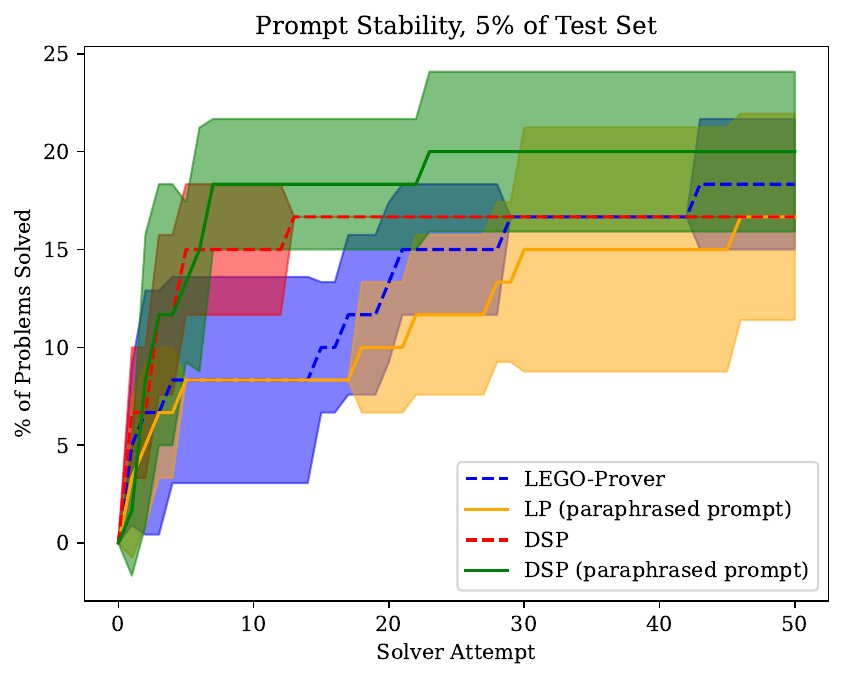}
    \caption{For \lp\ and \dsp\ we compare 5 runs with our default prompt versus 5 runs with different paraphrased prompts. We evaluate on a subset of the test set. We find that these system are fairly robust to minor changes in wording -- performance is typically within 1 standard deviation. These results are for a 5\% subset of the test split.}
    \label{fig:stab-test}
\end{figure}

We find no evidence that \lp\ is disadvantaged by prompt: accuracies are within one standard deviation, and there was no maximum potential gain (i.e., no problems were solved under one of the paraphrases, but not by the original prompt). See Table \ref{tab:stability} for details. Indeed, \dsp\ exhibited potential gain, suggesting that our \dsp\ prompt may be supoptimal. We also plot the performance per attempt in Figures \ref{fig:stab-val}
 and \ref{fig:stab-test}.

Note that we modify \cite{wahle-etal-2024-paraphrase}'s maximum potential gain metric to reflect that we collect data from 5 baseline trials instead of one. In our implementation, we consider it to be a 0\% improvement for a specific problem if a paraphrase achieves the same performance or worse (i.e., no improvement is possible for problems solvable under the baseline prompt), and 100\% if it solves a task that was unsolved under the baseline prompt. Averaging over all problems this becomes the fraction of problems that are only solved by a paraphrased prompt. The maximum potential loss is calculated in the same way but with the rewards reversed, thus becoming the fraction of problems that are solved by the baseline prompt, but by none of the paraphrased prompts. %

\raggedbottom

\section{Changes to Prompts} \label{sec:prompts}

\subsection{\dsp}

\dsp\ was originally designed for text completion models, specifically  Codex \texttt{code-davinci-002} \citep{DBLP:conf/nips/HendrycksBKMAGB21}. As it was not designed for conversational models, and none of the models evaluated support a text completion mode, we developed our own minimal system prompts to preface the few-shot examples preprended by \dsp\ to the task statement.

\subsubsection{gpt-4o-mini \& gpt-4o System Prompt}

In the prompt we emphasized the use of sledgehammer as it was a focus of the original \dsp\ paper, and the model would avoid its use unless told otherwise. After designing the prompt for gpt-4o-mini, we found that gpt-4o displayed no obviously incorrect behaviours, so we did not modify it further. We validated our prompts on a subset of two validation set problems (\texttt{mathd\_algebra\_109} and \texttt{algebra\_sqineq\_2at2pclta2c2p41pc}).

\definecolor{lightgreen}{rgb}{0.6,0.9,0.6}
\sethlcolor{lightgreen}

\begin{nohyphens}
    \noindent\fbox{%
        \parbox{\textwidth}{%
            As a mathematician familiar with Isabelle, your task is to complete the proof sketch provided by the user, MAKING SURE TO **ALWAYS** USE `sledgehammer` TO PROVE THE INTERMEDIATE RESULTS.
        }%
    }
\end{nohyphens}

\pagebreak 

\subsubsection{o3-mini System Prompt}

For o3-mini we modified the prompt to explicitly prohibit o3-mini from answering the in-context examples. We also removed the instruction to use sledgehammer as o3-mini would ignore the examples and attempt to use ``\texttt{by sledgehammer}'' instead of ``\texttt{sledgehammer}'' -- this would cause parsing errors in the human proof-correcting heuristics. Based on our small validation set of two problems, we found that removing the instruction to use sledgehammer worked more reliably than adding instructions to use sledgehammer correctly.

\definecolor{lightgreen}{rgb}{0.6,0.9,0.6}
\sethlcolor{lightgreen}

\begin{nohyphens}
    \noindent\fbox{%
        \parbox{\textwidth}{%
            Complete the Isabelle proof sketch provided by the user. The user will provide several examples, but you should only complete the last example (i.e., the example that does not have a proof).
        }%
    }
\end{nohyphens}

\subsection{\lp}

\lp\ was designed for conversational models, and as such has default system prompts for the \prover\ decomposer, \prover\ formalizer, and several \evolver\ variants. As a general rule, we attempted to minimize changes to these prompts where possible, doing so only when we found erroneous behaviour. \hl{Text in green} is added to the prompt, \hlrm{text in red} is removed from the original prompt.

\subsubsection{gpt-4o-mini prompts}

For gpt-4o-mini we found it necessary to add additional formatting instructions to the \prover's decomposer prompt:

\begin{nohyphens}
    \noindent\fbox{%
        \parbox{\textwidth}{%
As an mathematician and expert in isabelle theorem prover, your task is to analyze the given theorem (including problem's informal statement, human written informal proof, and formal statement). Provide a better structured step by step proof that closer to isabelle. and request relevant lemmas, theorems that might help in proving this problem.\\
\hl{You will format your answer as follows:}\\
\hl{- A section with the header `\#\# Structured informal proof` containing the  step-by-step proof.}\\
\hl{- A subsequent section with the header `\#\# Request skills` containing any requested relevant lemmas or theorems}\\
\hl{- Each requested skill is represented by an isabelle code block containing its formal definition}\\
        }%
    }
\end{nohyphens}

We also observed that the rate of rejection of proof attempts was high due to slight paraphrasing of the requested theorem statement. We addressed this by modifying the \prover's formalizer prompt:

\begin{nohyphens}
    \noindent\fbox{%
        \parbox{\textwidth}{%

As a mathematician familiar with Isabelle, your task is to provide a formal proof in response to a given problem statement.\\
Your proof should be structured and clearly written, meeting the following criteria:\\
- It can be verified by Isabelle.\\
- Each step of the proof should be explained in detail using comments enclosed in "(*" and "*)".\\
- The explanation for each step should be clear and concise, avoiding any unnecessary or apologetic language.\\
- You are **strongly encouraged** to create useful and reusable lemmas to solve the problem.\\
- The lemmas should be as general as possible (generalizable), and be able to cover a large step in proofs (non-trivial).\\
\hl{- You **MUST** copy the input lemma EXACTLY: **without changing whitespace**, without paraphrasing, and without adding comments.} \\
Please ensure that your proof is well-organized and easy to follow, with each step building upon the previous one.

        }%
    }
\end{nohyphens}

The \evolver\ prompts were unchanged.

\pagebreak 
\subsubsection{gpt-4o prompts}

For gpt-4o we used the gpt-4o-mini prompts and further modified the \prover's decomposer prompt to prevent it from decomposing the provided in-context examples.

\begin{nohyphens}
    \noindent\fbox{%
        \parbox{\textwidth}{%
            As an mathematician and expert in isabelle theorem prover, your task is to analyze the given theorem (including problem's informal statement, human written informal proof, and formal statement). Provide a better structured step by step proof that \hl{is} closer to isabelle. \hl{You will be given several examples; you must only decompose the last theorem given (i.e., the only theorem without a provided solution).}\\
\hl{Furthermore, you must} \hlrm{and} request relevant lemmas, theorems that might help in proving this problem.\\
\hl{You will format your answer as follows:}\\
\hl{- A section with the header `\#\# Structured informal proof` containing the  step-by-step proof.}\\
\hl{- A subsequent section with the header `\#\# Request skills` containing any requested relevant lemmas or theorems}\\
\hl{- Each requested skill is represented by an isabelle code block containing its formal definition}\\

        }%
    }
\end{nohyphens}

\subsubsection{o3-mini prompts}

For o3-mini we found it necessary to change all prompts to prevent the model from solving the in-context examples. We also found o3-mini struggled with placing the solution to proofs inside of Markdown code blocks, even after Markdown formatting was enabled by prepending ``Formatting re-enabled'' to the prompts.

For \prover's decomposer prompt we re-enabled Markdown formatting and added instructions to not decompose the in-context examples.

\begin{nohyphens}
    \noindent\fbox{%
        \parbox{\textwidth}{%
            \hl{Formatting re-enabled}\\
            As an mathematician and expert in isabelle theorem prover, your task is to analyze the given theorem (including problem's informal statement, human written informal proof, and formal statement). Provide a better structured step by step proof that closer to isabelle. and request relevant lemmas, theorems that might help in proving this problem. \hl{Note that you will be provided with several examples, but you should only complete the last example (i.e., the example that does not have a structured step by step proof and requests for relevant lemmas).}\\
            \hl{You will format your answer as follows:}\\
            \hl{- A section with the header `\#\# Structured informal proof` containing the  step-by-step proof.}\\
            \hl{- A subsequent section with the header `\#\# Request skills` containing any requested relevant lemmas or theorems}\\
            \hl{- Each requested skill is represented by an isabelle code block containing its formal definition}\\
            
        }%
    }
\end{nohyphens}

\pagebreak 
The \prover's formalizer prompt was modified to re-enable Markdown formatting and to emphasize the desired output format.

\begin{nohyphens}
    \noindent\fbox{%
        \parbox{\textwidth}{%
            \hl{Formatting re-enabled}\\
            As a mathematician familiar with Isabelle, your task is to provide a formal proof in response to a given problem statement.\\
            Your proof should be structured and clearly written, meeting the following criteria:\\
            - It can be verified by Isabelle.\\
            - Each step of the proof should be explained in detail using comments enclosed in "(*" and "*)".\\
            - The explanation for each step should be clear and concise, avoiding any unnecessary or apologetic language.\\
            - You are **strongly encouraged** to create useful and reusable lemmas to solve the problem.\\
            - The lemmas should be as general as possible (generalizable), and be able to cover a large step in proofs (non-trivial).\\
            \hl{- Your solution should be written within a Markdown isabelle code block. i.e., within a triple back ticked code block, labelled as "isabelle"}\\
            \hl{- You **MUST** copy the input lemma EXACTLY: **without changing whitespace**, without paraphrasing, and without adding comments.}
            \hl{- Within the code block, you must create a new Isabelle `theory` containing the aforementioned input lemma copied from the input, as well as your proof and any lemmas you created for the proof.}
            Please ensure that your proof is well-organized and easy to follow, with each step building upon the previous one.
        }%
    }
\end{nohyphens}

The evolver prompts were likewise modified. The do\_request prompt (immediately below) was particularly problematic as o3-mini tended to disregard the formatting instructions:

\begin{nohyphens}
    \noindent\fbox{%
        \parbox{\textwidth}{%
            \hl{Formatting re-enabled}\\
            As a mathematician familiar with Isabelle, your task is to provide a formal proof in response to a given formal statement.\\
            Your proof should be structured and clearly written, meeting the following criteria:\\
            - It can be verified by Isabelle.\\
            - Please ensure that your proof is well-organized and easy to follow, with each step building upon the previous one.\\
            \hl{- THIS IS **CRITICAL**: YOUR PROOF **MUST** BE WRITTEN WITHIN A MARKDOWN ISABELLE CODE BLOCK!!! i.e., within a triple back ticked code block, labelled as "isabelle"}\\
            \\
            \hl{Note that you will be provided with several examples, but you should only complete the last example (i.e., the example that does not have a formal proof).}\\
            \\
            \{examples\}\\
            \\
            \#\#\#\#\#\#\#\#\#\#\#\#\#\#\#\#\#\#\#\#\\
            \\
            \# Statement:\\
            ```isabelle\\
            \{formal\_statement\}\\
            ```\\
            \\
            \# Proof\\

        }%
    }
\end{nohyphens}

\begin{nohyphens}
    \noindent\fbox{%
        \parbox{\textwidth}{%
            \hl{Formatting re-enabled}\\
            As an expert mathematician who is proficient in Isabelle theorem proving, your task is to modify the given lemma, theorem, function or definition given in the code to aid in solving one or more of the problems provided. Your should accomplish this by Extend Dimensions: If the problem is defined in a specific number of dimensions, consider if it would still hold in more or fewer dimensions.\\
            \hl{Your answer **MUST** be written within a Markdown isabelle code block. i.e., within a triple back ticked code block, labelled as "isabelle"}\\
            \\
            Here is some reference problems, you should evolve the skill to help solving theses problems:\\
            \{problems\}\\
            \\
            \hl{Note that you will be provided with several examples, but you should only complete the last example (i.e., the example that does not have a modified version).}\\
            \\
            \#\#\#\#\#\#\#\#\#\#\#\# Extend Dimensions \#\#\#\#\#\#\#\#\#\#\#\#\\
            \{examples\}\\
            \\
            \\
            \#\#\#\#\#\#\#\#\#\#\#\#\#\#\#\#\#\#\#\#\\
            \#\# Skill to evolve\\
            ```isabelle\\
            \{code\}\\
            ```\\
            \\
            \#\# Evolved skill\\
            
        }%
    }
\end{nohyphens}

\begin{nohyphens}
    \noindent\fbox{%
        \parbox{\textwidth}{%
            \hl{Formatting re-enabled}\\
            As an expert mathematician who is proficient in Isabelle theorem proving, your task is to modify the given lemma, theorem, function or definition given in the code to aid in solving one or more of the problems provided. Your should accomplish this by `Identifying Key Concepts`: Extract the essential ideas, methods, or theorems that are critical to solving the problem.\\
            \hl{Your answer **MUST** be written within a Markdown isabelle code block. i.e., within a triple back ticked code block, labelled as "isabelle"}\\
            \\
            Here is some reference problems, you should evolve the skill to help solving theses problems:\\
            \{problems\}\\
            \\
            \hl{Note that you will be provided with several examples, but you should only complete the last example (i.e., the example that does not have a modified version).}\\
            \\
            \#\#\#\#\#\#\#\#\#\#\#\# Identifying key Concepts \#\#\#\#\#\#\#\#\#\#\#\#\\
            \{examples\}\\
            \\
            \\
            \#\#\#\#\#\#\#\#\#\#\#\#\#\#\#\#\#\#\#\#\\
            \#\# Skill to evolve\\
            ```isabelle\\
            \{code\}\\
            ```\\
            \\
            \#\# Evolved skill\\

        }%
    }
\end{nohyphens}

\begin{nohyphens}
    \noindent\fbox{%
        \parbox{\textwidth}{%
            \hl{Formatting re-enabled}\\
            As an expert mathematician who is proficient in Isabelle theorem proving, your task is to modify the given lemma, theorem, function or definition given in the code to aid in solving one or more of the problems provided. Your should accomplish this by Parameterize: If the problem involves specific numbers, generalize it by replacing these with variables.\\
            \hl{Your answer **MUST** be written within a Markdown isabelle code block. i.e., within a triple back ticked code block, labelled as "isabelle"}\\
            \\
            Here is some reference problems, you should evolve the skill to help solving theses problems:\\
            \{problems\}\\
            \\
            \hl{Note that you will be provided with several examples, but you should only complete the last example (i.e., the example that does not have a modified version).}\\
            \\
            \#\#\#\#\#\#\#\#\#\#\#\# Parameterize \#\#\#\#\#\#\#\#\#\#\#\#\\
            \{examples\}\\
            \\
            \\
            \#\#\#\#\#\#\#\#\#\#\#\#\#\#\#\#\#\#\#\#\\
            \#\# Skill to evolve\\
            ```isabelle\\
            \{code\}\\
            ```\\
            \\
            \#\# Evolved skill\\

        }%
    }
\end{nohyphens}

\begin{nohyphens}
    \noindent\fbox{%
        \parbox{\textwidth}{%
            \hl{Formatting re-enabled}\\
            As an expert mathematician who is proficient in Isabelle theorem proving, your task is to modify the given lemma, theorem, function or definition given in the code to aid in solving one or more of the problems provided. Your should accomplish this by Scale Complexity: Try both simpler and more complicated versions of the problem to see how the approach adapts.\\
            \hl{Your answer **MUST** be written within a Markdown isabelle code block. i.e., within a triple back ticked code block, labelled as "isabelle"}\\
            \\
            Here is some reference problems, you should evolve the skill to help solving theses problems:\\
            \{problems\}\\
            \\
            \hl{Note that you will be provided with several examples, but you should only complete the last example (i.e., the example that does not have a modified version).}\\
            \\
            \#\#\#\#\#\#\#\#\#\#\#\# Scale Complexity \#\#\#\#\#\#\#\#\#\#\#\#\\
            \{examples\}\\
            \\
            \\
            \#\#\#\#\#\#\#\#\#\#\#\#\#\#\#\#\#\#\#\#\\
            \#\# Skill to evolve\\
            ```isabelle\\
            \{code\}\\
            ```\\
            \\
            \#\# Evolved skill\\

        }%
    }
\end{nohyphens}

\section{Calculation of Cost} \label{app:cost_plot_creation}

In our experiments we recorded the total token usage per run. I.e., if an experiment run consisted of 100 attempts, then we recorded the cost of 100 attempts. However, to create Table \ref{tab:eval_by_cost} we required the cost for a reduced number of attempts. To estimate this, we determined the average cost per problem attempt. I.e., we calculated the total cost over all runs and divided this by the total number of problem attempts over all runs. For example, the 3 runs of the gpt-4o-mini experiment on the full test set used 100 attempts per problem each; therefore, we divided the total cost of these 3 experiment runs by 300 to determine the average cost per attempt.

Similarly, Figure \ref{fig:cost} required the total expenditures up to the $n$th success, and we again estimated this using the average cost per attempt. Specifically, we plotted $(x,y)$ points where $x$ was the cost of $i$ attempts on all remaining problems, and $y$ was the number of problems solved after $i$ attempts.

For further details, the code used to calculate Table \ref{tab:eval_by_cost} and plot Figure \ref{fig:cost} is publicly available in our GitHub repository.

\section{Soft Use Score Examples}\label{app:soft-use-examples}

In this section we report examples of various retrieved lemmas and their soft use scores. The  example chosen (\texttt{data/full\_data/test/amc12a\_2021\_p9.json}) was generated in the first run of gpt-4o-mini; it was selected as it has the largest range of soft use values.

The English description for this task is:

\begin{nohyphens}
    \noindent\fbox{%
        \parbox{\textwidth}{%
Which of the following is equivalent to
$(2+3)(2^2+3^2)(2^4+3^4)(2^8+3^8)(2^{16}+3^{16})(2^{32}+3^{32})(2^{64}+3^{64})?$

$\textbf{(A)} ~3^{127} + 2^{127} \qquad\textbf{(B)} ~3^{127} + 2^{127} + 2 \cdot 3^{63} + 3 \cdot 2^{63} \qquad\textbf{(C)} ~3^{128}-2^{128} \qquad\textbf{(D)} ~3^{128} + 2^{128} $ 

Show that it is $\textbf{(C)} ~3^{128}-2^{128}$.

        }%
    }
\end{nohyphens}

\pagebreak 

The model output prior to correction by heuristics is reproduced below. Some lines have been word wrapped to fit within margins. 

\begin{Verbatim}[frame=single]
theory amc12a_2021_p9 
  imports Complex_Main 
begin

lemma prod_sum_of_powers:
  fixes n :: nat
  shows "(\<Prod> k < n. (2^(2^k) + 3^(2^k))) = (3::nat)^(2^n) - 2^(2^n)"
proof (induction n)
  case 0
  then show ?case 
    by auto
next
  case (Suc n)
  then have IH: "(\<Prod> k < n. (2^(2^k) + 3^(2^k))) = 
                                                   (3::nat)^(2^n) - 2^(2^n)" 
    by simp
  have "(\<Prod> k < Suc n. (2^(2^k) + 3^(2^k))) = 
                  (\<Prod> k < n. (2^(2^k) + 3^(2^k))) * (2^(2^n) + 3^(2^n))"
    by (simp add: prod.atLeastLessThan_Suc)
  also have "... = ((3::nat)^(2^n) - 2^(2^n)) * (2^(2^n) + 3^(2^n))" 
    using IH by simp
  also have "... = 3^(2^n) * 3^(2^n) - 2^(2^n) * 2^(2^n)" 
    by (simp add: algebra_simps)
  also have "... = 3^(2^(Suc n)) - 2^(2^(Suc n))" 
    by (metis power2_eq_square power_Suc power_even_eq semiring_norm(3))
  finally show ?case by simp
qed

theorem amc12a_2021_p9 :
  shows "(\<Prod> k<7. (2^(2^k) + 3^(2^k))) = (3::nat)^128 - 2^128"
proof -
  have "(\<Prod> k<7. (2^(2^k) + 3^(2^k))) = 
                                   (\<Prod> k<7. (2^(2^k) + 3^(2^k)))" by simp
  then have "(\<Prod> k<7. (2^(2^k) + 3^(2^k))) = (3::nat)^(2^7) - 2^(2^7)"
    using prod_sum_of_powers[of 7] by simp
  then have "(\<Prod> k<7. (2^(2^k) + 3^(2^k))) = 
                                    (3::nat)^(128) - 2^(128)" by simp
  thus ?thesis by simp
qed

end
\end{Verbatim}

\pagebreak 
The following retrieved lemma provided to the \prover\ when generating the solution above attained a soft use score of $0.378$. It is largely irrelevant, as it is specialized to a function that does not appear in this problem.

\begin{Verbatim}[frame=single]
definition f :: "nat \<Rightarrow> nat" where
  "f x = 4^x + 6^x + 9^x"

lemma f_12k_plus_9:
  fixes k :: nat
  shows "f (12 * k + 9) = 4^(12 * k + 9) + 6^(12 * k + 9) + 9^(12 * k + 9)"
proof -
  have "f (12 * k + 9) = 4^(12 * k + 9) + 6^(12 * k + 9) + 9^(12 * k + 9)" 
    by (simp add: f_def)
  thus ?thesis by simp
qed
\end{Verbatim}

The following retrieved lemmas provided to the \prover\ when generating the solution above attained soft use scores of $0.695$ and $0.706$ respectively. They are more relevant, involving powers of powers, but are again overly specialized and the proof tactics used (\texttt{simp add: power\_add}, \texttt{simp add: power2\_eq\_square}, etc...) are not used in the \prover's outputted solution. Note that the \lp's \evolver\ often produces minor variations of the same lemma as seen in this case.

\begin{Verbatim}[frame=single]
lemma power_computation:
  fixes k :: nat
  shows "k^12 = (k^6)^2"
proof -
  have "k^12 = k^(6 + 6)" by simp
  also have "... = k^6 * k^6" by (simp add: power_add)
  also have "... = (k^6)^2" by (simp add: power2_eq_square)
  finally show ?thesis by simp
qed
\end{Verbatim}

\begin{Verbatim}[frame=single]
lemma power_computation:
  fixes k :: nat
  shows "k^12 = (k^6)^2"
proof -
  have "k^12 = k^(6 + 6)" by simp
  also have "... = (k^6)^2" by (simp add: power_add)
  finally show ?thesis .
qed
\end{Verbatim}

\pagebreak
The following retrieved lemma provided to the \prover\ when generating the solution above attained a soft use score of $0.96$. Note that some lines are word wrapped to fit within the margins. It is reproduced almost exactly in the \prover's output, with minor changes to parenthesization and the name of the lemma. 

\begin{Verbatim}[frame=single]
lemma sum_of_powers:
  fixes n :: nat
  shows "(\<Prod> k < n. (2^(2^k) + 3^(2^k))) = (3::nat)^(2^n) - 2^(2^n)"
proof (induction n)
  case 0
  then show ?case 
    by auto
next
  case (Suc n)
  then have IH: "(\<Prod> k < n. (2^(2^k) + 3^(2^k))) = 
                                               (3::nat)^(2^n) - 2^(2^n)" 
    by simp
  have "(\<Prod> k < Suc n. (2^(2^k) + 3^(2^k))) = 
              (\<Prod> k < n. (2^(2^k) + 3^(2^k))) * (2^(2^n) + 3^(2^n))"
    by (simp add: prod.atLeastLessThan_Suc) 
  also have "... = ((3::nat)^(2^n) - 2^(2^n)) * (2^(2^n) + 3^(2^n))" 
    using IH by simp
  also have "... = (3^(2^n) * 3^(2^n)) - (2^(2^n) * 2^(2^n))" 
    by (simp add: algebra_simps)
  also have "... = 3^(2^(Suc n)) - 2^(2^(Suc n))" 
    by (metis power2_eq_square power_Suc power_even_eq semiring_norm(3))
  finally show ?case by simp
qed
\end{Verbatim}

\end{document}